\documentclass{ifacconf}
\makeatletter
\let\old@ssect\@ssect 
\makeatother
\usepackage{ifpdf}
\ifpdf
	\usepackage[hidelinks]{hyperref}
\fi
\usepackage[T1]{fontenc} 
\usepackage{selinput}\SelectInputMappings{adieresis={ä},germandbls={ß}}
\usepackage{scrhack} 
\usepackage{lmodern}
\usepackage[babel]{microtype} 
\usepackage{textcomp} 

\usepackage{xcolor} 

\usepackage{amsmath}
\usepackage{amssymb}
\usepackage{calc}
\usepackage{amstext} 
\usepackage{natbib}        
\usepackage{algorithm}
\usepackage[noend]{algpseudocode}
	\makeatletter
	\def\BState{\State\hskip-\ALG@thistlm}
	\makeatother
\makeatletter
\newcommand{\StatexIndent}[1][3]{%
	\setlength\@tempdima{\algorithmicindent}%
	\Statex\hskip\dimexpr#1\@tempdima\relax}

\usepackage{blindtext}

\usepackage{nicefrac} 
\usepackage{units} 

\usepackage[super]{nth}

\usepackage{pstricks}
\usepackage{graphicx}
\usepackage{import}
\usepackage{calc}
\usepackage[crop=pdfcrop]{pstool} 

\usepackage{tabularx} 
\usepackage{booktabs}
\usepackage{makecell} 

\usepackage[babel,german=quotes]{csquotes} 

\newcommand{\hilodoe}{HiLoMoT-DoE}

\newcommand{\NRMSEF}{{\text{NRMSE}}}

\newcommand{\NRMSEValF}{\NRMSEF_{\text{val}}} 
\newcommand{\NRMSEVal}{$\NRMSEValF{}$}

\newcommand{\CVF}{{\text{CV}}_{K}} 
\newcommand{\CV}{$\CVF{}$}
\newcommand{\CVtenF}{{\text{CV}}_{10}} 
\newcommand{\CVten}{$\CVtenF{}$}
\newcommand{\CVhighF}{{\text{CV}}_{10\text{, high}}} 
\newcommand{\CVhigh}{$\CVhighF{}$}
\newcommand{\CVnrmF}{{\CVtenF}\!_{\text{, nrm}}} 
\newcommand{\CVnrm}{$\CVnrmF{}$}

\newcommand{\nMeasF}{{n}_{\text{Meas}}} 
\newcommand{\nMeas}{$\nMeasF{}$}

\newcommand{\SNRF}{\textrm{SNR}}
\newcommand{\SNR}{$\SNRF{}$}

\newcommand{\mLF}{m_{\textrm{L}}}
\newcommand{\mL}{$\mLF{}$}

\newcommand{\ofM}{\!_{, m}}

\DeclareMathOperator{\yHat}{\mathnormal{\hat{y}}}
\DeclareMathOperator{\varHat}{{\hat{\sigma}}^{2}}
\DeclareMathOperator{\varHatM}{{\hat{\sigma}}^{2}_{\mathnormal{m}}}
\DeclareMathOperator{\varN}{{\sigma}^{2}_{\mathnormal{n}}}
\DeclareMathOperator{\stdN}{{\sigma}_{\mathnormal{n}}}
\newcommand{\xHat}{\hat{x}}
\newcommand{\xAst}{x^{\ast}}
\newcommand{\xAstM}{x^{\ast}_{m}}
\newcommand{\ofX}{(\xHat)}
\newcommand{\ofXAstM}{(\xAstM)}
\newcommand{\ofXX}{(\xHat,\xHat)}
\DeclareMathOperator{\kHat}{\hat{\text{\textbf{k}}}}
\DeclareMathOperator{\covOp}{\mathnormal{k}}

\newcommand{\yHatOfX}{\yHat\ofX}
\newcommand{\varHatOfX}{\varHat\ofX}
\newcommand{\varMOfXAst}{\varHatM\ofXAstM}
\newcommand{\covWithNoise}{(K+\varN I)}
\newcommand{\covOfXX}{\covOp\ofXX}
\newcommand{\yMatrix}{\text{\textbf{y}}}

\DeclareMathOperator*{\argmax}{arg\,max}

\newcommand{\isep}{\mathrel{{.}\,{.}}\nobreak}

\makeatletter
\def\@ssect#1#2#3#4#5#6{%
	\NR@gettitle{#6}
	\old@ssect{#1}{#2}{#3}{#4}{#5}{#6}
}
\makeatother

\begin{document}
\begin{frontmatter}

\title{Improved active output selection strategy for noisy environments}
%
\author[First]{Adrian~Prochaska} 
\author[First]{Julien~Pillas} 
\author[Second]{Bernard~Bäker}

\address[First]{Mercedes-Benz AG,
	71059 Sindelfingen,
	Germany
}
\address[Second]{Institute of Automotive Technology Dresden,
	George-Bähr-Str. 1b,
	01062 Dresden,
	Germany
}

\begin{abstract}                
	The test bench time needed for model-based calibration can be reduced  with active learning methods for test design.
	This paper presents an improved strategy for active output selection.
	This is the task of learning multiple models in the same input dimensions and suits the needs of calibration tasks.
	Compared to an existing strategy, we take into account the noise estimate, which is inherent to Gaussian processes.
	The method is validated on three different toy examples.
	The performance compared to the existing best strategy is the same or better in each example.
	In a best case scenario, the new strategy needs at least 10\% less measurements compared to all other active or passive strategies.
	Further efforts will evaluate the strategy on a real-world application.
	Moreover, the implementation of more sophisticated active-learning strategies for the query placement will be realized.
\end{abstract}

\begin{keyword}
Gaussian Processes, Active Learning, Regression, Active Output Selection, Drivability Calibration, Experiment Design
\end{keyword}

\end{frontmatter}

\section{Introduction}
\label{sec:introduction}

Active learning -- also known as \textit{online design of experiments} or \textit{optimal experimental design} -- reduces the number of necessary measurements to achieve sufficient model accuracy.
This is especially relevant for tasks with high labeling or measurement costs.
Additionally, it facilitates the test design procedure since the placement of measurement points is conducted during test execution.
The scientific community mostly focuses on optimally learning regression models with one output.
Our field of research and application, the drivability calibration, requires models for multiple outputs with the same input dimensions.
There is an importance of assuring a balanced model accuracy for all outputs, which originates in their purpose: 
The identified models are mostly used for optimization. 
In this case, the model with the lowest accuracy determines the quality of the following optimization.
For a test engineer, this raises a question: 
\textit{In which order should the models actively learn to have the lowest number of measurements?}

For these boundary conditions, we introduced a new active learning task, which is called \textit{active output selection}, in a previous study, see \cite{prochaska_active_2020}.
We compared already existing heuristic strategies to a new strategy which chooses the leading model based on the cross-validation error.
This paper advances the presented strategy to overcome the drawbacks.
Multiple toy examples are used to illustrate the effectiveness of the new strategy.

Section~\ref{sec:previousWorks} of this paper introduces previous works in context of active learning in general and in particular for regression tasks. 
A special focus lies on applications in the automotive calibration domain.
Section~\ref{sec:problemDefinition} focuses on the task of \textit{active output selection} and the specialties of active learning in the context of drivability calibration. 
Section~\ref{sec:strategies} describes the analyzed approaches. 
Furthermore, an improved strategy for active output selection is presented. 
The approaches are evaluated using different toy examples.
Experimental details and a discussion of results are shown in section~\ref{sec:experiments}.
At the end, section~\ref{sec:conclusion} concludes the results and presents fields of possible future works.

\section{Previous works}
\label{sec:previousWorks}

The field of active learning is a growing field in machine learning.
In statistics literature, it is also called \textit{optimal experimental design}, see \cite{cohn_neural_1996}.
Since there is a necessity of labeled data, it is a subdomain of supervised learning.
\cite{settles_active_2009} illustrates a broad overview of the current state of the art in active learning and gives an outlook to multiple possible future work fields.
Recent advances in the scientific community mostly focused on classification problems. 
Those methods are often applied to speech recognition and text information extraction tasks, see \cite{settles_active_2009}.

Regression tasks in the context of active learning have not been as popular.
However, the methodological advances are relevant as well.
\cite{sugiyama_active_2008} showed an approach which actively learns an ensemble of models for the same task and selects the best one to query new points.
\cite{cai_maximizing_2013} introduced expected model change maximization (EMCM) for improving active learning for gradient boosted decision trees.
\cite{cai_batch_2017} later extended EMCM to choose a set of informative queries and to Gaussian process regression models (GPs).
\cite{park_robust_2020} evolved EMCM for active learning even further into a learning algorithm, which handles outliers more robustly than before.
The active learning strategies presented in those publications focus on improvements for single-output regression models. 
\cite{zhang_near-optimal_2016} introduced a learning algorithm for multiple-output Gaussian processes (MOGP).
The algorithm improves the prediction accuracy of one target output model by taking into account several correlated auxiliary output models. 
It outperforms multiple single-output Gaussian processes (SOGP) and additionally indicates that a global consideration of multiple outputs is beneficial.

In automotive calibration tasks, the identification of multiple process outputs in the same experiment is more relevant to the application.
There were also advances in active learning in this field.
\cite{hartmann_adaptive_2013} presented \hilodoe{}, a method for design of experiments with hierarchical local model trees.
\cite{klein_adaptive_2013} applied this method successfully to an engine calibration task.
Each of their two application examples had two outputs as well as five an seven inputs, respectively.
A sequential strategy was chosen to model the two outputs in the application example.
This strategy defines one output model as leading and identifies it completely before moving to another output.

\cite{reichart_multi-task_2008} introduced a round-robin strategy for combining two linguistic annotation tasks\footnote{The round-robin strategy is called alternating selection by \cite{reichart_multi-task_2008}.}.
It switches the leading model after each iteration and performs better than separate annotation.

\cite{dursun_ansatz_2015} adapted \hilodoe{} to drivability calibration examples.
Their algorithm identified multiple static regression tasks with identical input dimensions during the test run.
Furthermore, they applied the round-robin strategy of \cite{reichart_multi-task_2008} to their regression task and compared it to other existing strategies.
In this single real-world experiment, the round-robin strategy performs better than offline methods and the online sequential strategy. 
This indicates, that round-robin is preferably used in general.

In an earlier publication, we thoroughly analyzed those strategies in a reproducible, noisy environment, see \cite{prochaska_active_2020}.
We introduced a new strategy, which selects the leading model based on the evaluation of the cross-validation error.
The results can be described as mixed.
Depending on the setup, it outperforms all other strategies by up to 20\%.
However, the strategy has difficulties when dealing with different signal-to-noise-ratios on different outputs.
This leads to the conclusion, that the strategy's general idea is advantageous, but its robustness to noise has to be increased.

\section{Problem definition}
\label{sec:problemDefinition}
In general, active learning is applied to tasks with high labeling costs.
As stated in section~\ref{sec:previousWorks}, most advances in active learning improve the learning strategies for one model output.
This paper is a contribution to a subdomain of active learning which is called \textit{active output selection} (AOS).
For these tasks, an algorithm identifies $M>1$ models.
Their outputs are possibly uncorrelated, but their input dimensions are identical.
In case of drivability calibration, measurements determine the values of the modeled outputs.
The measurements always contain noise to some extent.
A measure for the noise of output $m$ is defined by the signal-to-noise-ratio $\SNRF\!_{m}=\frac{\max\left({y_{m}}\right)-\min\left({y_{m}}\right)}{\sigma_{\textrm{N}}}$, where $y_{m}$ is the vector of all measurements of output $m$ and $\sigma_{\textrm{N}}$ the standard deviation of normally distributed noise.
For drivability criteria, the \SNR{} lies approximately in a range of $\left(7\isep{} 100\right)$ and can be different for each output.
The measurements are conducted on powertrain test benches.
On the one hand, one measurement is cost-expensive, hence the number of measurements ought to be reduced by active learning.
Furthermore, the costs of measuring one output are identical to measuring all outputs, which then results in always measuring all $M$ outputs for one query.
On the other hand, the duration of one measurement is timely more expensive than the evaluation of code.
Consequently, a specific complexity analysis of the presented algorithms is lacking in this paper, but advantages and disadvantages are discussed in section~\ref{sec:strategies}.
After identification, the models are used for optimizations of control unit parameters.
Therefore, there is no primary or target output, but an adequate model quality for all $M$ models is crucial.

\section{Active output selection strategies}
\label{sec:strategies}
Each of the $M$ process outputs is modeled using GPs, since they are suited for drivability applications and handle noise robustly, see \cite{tietze_model-based_2015}.
In each iteration of the active learning setup, the leading process output defines the query for the next measurement.
This paper analyzes different strategies to determine the leading process output. 
The queries are placed by the leading output using a maximum variance strategy, which was presented by \cite{mackay_information-based_1992} in a general form.
The next query $\xAstM$ of output $m$, which is inside the input space $\mathbb{X}$, maximizes the same model's output variance $\varHatM$.
\begin{equation}
	\xAstM = \argmax_{\xAstM \in \mathbb{X}}{\left( \varMOfXAst \right) } \label{eq:maxVarianaceStrategy}
\end{equation}
The implementation is straightforward for GPs since the output variance is directly evaluated at each individual input point. 
Equation~\ref{eq:GPoutput} and eq.~\ref{eq:GPvariance} show the calculations of the predicted mean $\yHat$ and output variance $\varHat$ of a GP. 
$\kHat$ is the vector of covariances $\covOp(X,\xHat)$ between the measured training points $X$ and a single test point $\xHat$.
$K=K(X,X)$ are the covariances of $X$ and $\yMatrix$ contains the observations under noise with variance $\varN$, see \cite{rasmussen_gaussian_2008}.
\begin{align}
	\label{eq:GPoutput}
	\yHatOfX &= \kHat^{T} \, \covWithNoise^{-1} \, \yMatrix
	\\
	\label{eq:GPvariance}
	\varHatOfX &= \covOfXX - \kHat^{T} \covWithNoise^{-1} \kHat
\end{align}


The active learning strategy defines the next query if the leading model is already selected.
However, an AOS strategy has to define the leading model of the learning process beforehand.
In the following, all analyzed AOS strategies are described.
These descriptions include a space-filling, a round-robin and a \CVhigh{} strategy.
Furthermore, this paper presents a normalized \CVhigh{} strategy, which is an improvement of of the \CVhigh{} strategy.
To the author's knowledge, the latter has never been presented before and therefore is entirely new.

\subsubsection{sequential space-filling strategy (passive, SF)}
This is the baseline strategy. 
We do not assume any prior knowledge about the modeled process outputs. 
In drivability calibration, the usage of space-filling designs with a high number of points is common.
An offline space-filling design can only be applied if either the number of measurements is fixed before the experiments.
Contrary to that, the sequential space-filling strategy is passive but can be compared with other active methods, since the number of points is sequentially increased.
As all AOS strategies, SF starts with a number of random points. 
Every succeeding point $\xAst{}$ maximizes the minimum Mahalanobis-distance $d_{\min}{(\xHat)}=\min{\left\|{\xHat}-{X}\right\|}$ between the already measured points $\xHat{}$ and a huge set of candidate points $X$.\footnote{For uncorrelated data in a range between 0 and 1, the Mahalanobis distance and the Euclidean distance are the same.}
This strategy does not use active learning at all, since the models do not have an impact on the query placement.
In contrast to AOS strategies, SF is a passive strategy.

\subsubsection{round-robin strategy (RR)}
This is the second baseline strategy.
RR changes the leading model after each iteration.
When a model reaches the desired model quality, it is no leading model until the end of the experiment.
RR does not compare model qualities of the process outputs with each other.
This leads RR to not operating optimally because models that cannot be further improved but do not reach the desired model quality or models that do not benefit from being in the lead at the current status of experiment are still active every $M$ iterations.
Due to the characteristics of this heuristic strategy, RR is robust under noisy conditions and outperforms SF in general.

\subsubsection{\CVhigh{} strategy (CVH)}
This strategy is presented and analyzed in \cite{prochaska_active_2020}.
It assumes that the model with the lowest model accuracy benefits the most from leading the learning process.
The algorithm uses the normalized root mean squared $K$-fold cross-validation-error \CV{} with $K=10$.
\CVten{} is filtered with a digital moving average filter for stability reasons, see \cite{prochaska_active_2020}.
In each iteration, the CVH calculates \CVten{} for all outputs and selects the one with the highest \CVten{} as leading.
Each of the $M$ models has to be identified for $K$ times to calculate \CV{}, which results in higher computational costs compared to other strategies.

\cite{prochaska_active_2020} showed that CVH is beneficial in most setups. 
It handles similar and different complexities in model outputs well and outperforms baseline as well as other AOS strategies.
However, it is not able to distinguish between bad model quality and high noise.
Setups with an uncomplex but noisy output, on the one hand, and a complex but less noisy output, on the other hand, lead to poor performance.
Even the SF baseline outperforms CVH in such scenarios.
This lead to the conclusion that CVH has to be improved into handling such scenarios better.

\subsubsection{normalized \CVhigh{} strategy (CVHn)}
The drawbacks described in the previous paragraph show that an improvement of the existing AOS strategy is necessary.
CVHn picks up the idea of selecting a model based on its \CVten{}.
However, it includes a noise estimate in the decision criterion.
Therefore, CVHn chooses the leading model \mL{} based on the normalized \CVten{}.
This strategy uses a characteristic of the GPs.
As shown in equation~\ref{eq:GPvariance}, a standard deviation estimate of the process noise $\stdN$ is directly determined during the model identification. 
Since the real noise of a process output limits the minimum possible model error, \CVten{} is normalized with the noise estimate $\stdN{}$.
We assume that the model $\mLF{}=\argmax{\left( \CVnrmF{}\ofM{} \right)}$ with a high normalized cross-validation score $\CVnrmF{}\ofM{}=\frac{\CVtenF{}\ofM{}}{\stdN\ofM}$ benefits most from leading the learning process.

Algorithm~\ref{alg:CVHn} shows the pseudo-code of CVHn.
First, CVHn measures an initial set of points. 
Afterwards, the algorithm identifies all $M$ models in each iteration and calculates \CVnrm{}.
The algorithm defines the model that has the highest \CVnrm{} as leading.
CVHn stops if the maximum number of points or the desired model quality is reached. 

\begin{algorithm}[h]
	\caption{CVHn active output selection strategy.}\label{alg:CVHn}
	\begin{algorithmic}[1]
		\Repeat
		\If {no initial points have been carried out}
		\State plan queries of initial points
		\Else
		\State find model with the highest normalized 
		\StatexIndent[2] cross-validation error
		\State calculate next query
		\EndIf
		\State conduct measurements on planned queries
		\ForAll{models}
		\State update model
		\State assess cross-validation error
		\State filter the cross-validation error
		\State assess process noise estimate
		\State normalize cross-validation error
		\EndFor
		\Until{maximum number of points or desired model quality is reached}
	\end{algorithmic}
\end{algorithm}

As CVH, CVHn has to identify all $M$ models for $K$ times.
This leads to higher computational costs compared to the other strategies.
The execution time of those algorithms is not crucial for the drivability calibration applications on test benches, since the time needed to conduct one measurement is much bigger.

\section{Experiments}
\label{sec:experiments}

As described previously, all AOS strategies have different strengths and disadvantages.
The following section analyzes the effectiveness of those strategies on toy examples.
The AOS strategies are tested for three different toy examples.
These three examples are specially designed to show their characteristics.
15 runs are conducted for each AOS strategy and for each toy example.
Any random seeds are reset for each run of one strategy but changed after every round.
Due to the same prerequisites, one run of each strategy is comparable to another.
The model accuracy of each process output is evaluated with the \NRMSEVal{} (see eq.~\ref{eq:NRMSEVal}).
\begin{equation}
	\label{eq:NRMSEVal}
	\NRMSEValF\!_{,m}=\sqrt{\frac{\sum_{i=1}^{N_{\text{val}}}{\left ( {y}_{m,i\text{,val} } - {\hat{y}}_{m,i\text{,val}} \right)^{2}} }{\max\left({y_{m}}\right)-\min\left({y_{m}}\right)}}
\end{equation}
To compare multiple strategies with each other, one single criterion is favored over $M$ different ones.
Therefore, $\NRMSEValF{}\!_{\text{,}\varSigma}$, the squared sum of the accuracy criteria of all process outputs, is used.
\begin{equation}
	\label{eq:NRMSEValSum}
	\NRMSEF_{\text{val,}\varSigma} = \sqrt{\sum_{m=1}^{M}{\left( \NRMSEF_{\text{val,}m}\right) ^{2}}} 
\end{equation}

To test the algorithms, three different setups of toy examples that reflect distinct possible real world cases are used.
These setups are described in \cite{prochaska_active_2020} and only mentioned in this paper.
Each setup has three different outputs that are non-correlated.
Each of those outputs has an underlying analytical model, which is created by a random function generator, which was introduced by \cite{belz_proposal_2015}. 
The output values are overlaid with normally distributed noise; the standard deviation varies from setup to setup.
Each output has two input dimensions in the same range.
Because this toy example is derived from test bench applications, querying one input $x$ from the setup hands values for each output over to the learning algorithm (see section~\ref{sec:problemDefinition}).
The three setups are distinguished by the complexity of the underlying model as well as the signal-to-noise-ratio \SNR{} of each output:
\begin{itemize}
	\item Setup~1 has similar model complexities and \SNR{}s for all outputs.
	\item Setup~2 has similar \SNR{}s for all outputs but one output with considerably higher complexity. The complexity of said output is due to step functions, which represent a strong local non-linearity.
	\item Setup~3 also has one output with higher complexity. Additionally, a different output is characterized by a lower \SNR{}.
\end{itemize}
The number of measurements \nMeas{} is limited to 100 for each strategy and run.
For validation purposes, 121 points on an equidistant 11-by-11-grid are considered.
The advantage of the toy examples is that the true validation point without the influence of any output noise are known.

\begin{figure*}
	\centering
	\includegraphics{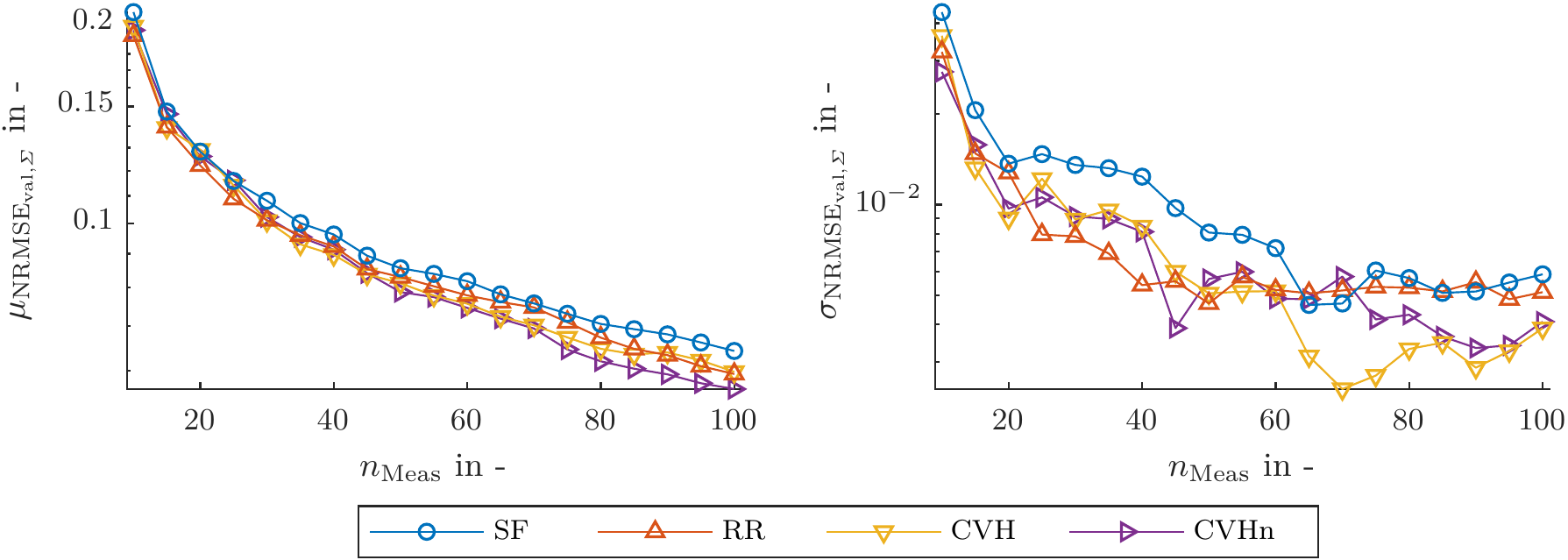}
	\caption{Mean $\mu$ and standard deviation $\sigma$ of the $\NRMSEF_{\text{val,}\varSigma}$ of setup~1 over the number of measurements \nMeas{}.}
	\label{fig:similarComplexity}	
\end{figure*}

Figure~\ref{fig:similarComplexity} shows the mean and standard deviation of the validation error over the number of measurements for setup~1.
The active learning strategies outperform the passive SF.
RR and CVH reduce the number of points to reach SF's end value by 10\% and 15\%, respectively.
CVHn further reduces the number of measurement points and reaches the end value of SF after 80\%.
For $\nMeasF\geq50$, CVHn performs significantly better than SF.
In this setup, CVHn outperforms the other active output selection strategies after $\nMeasF=90$.
The passive and all active learning strategies strike with low variance over the runs.

\begin{figure*}
	\centering
	\includegraphics{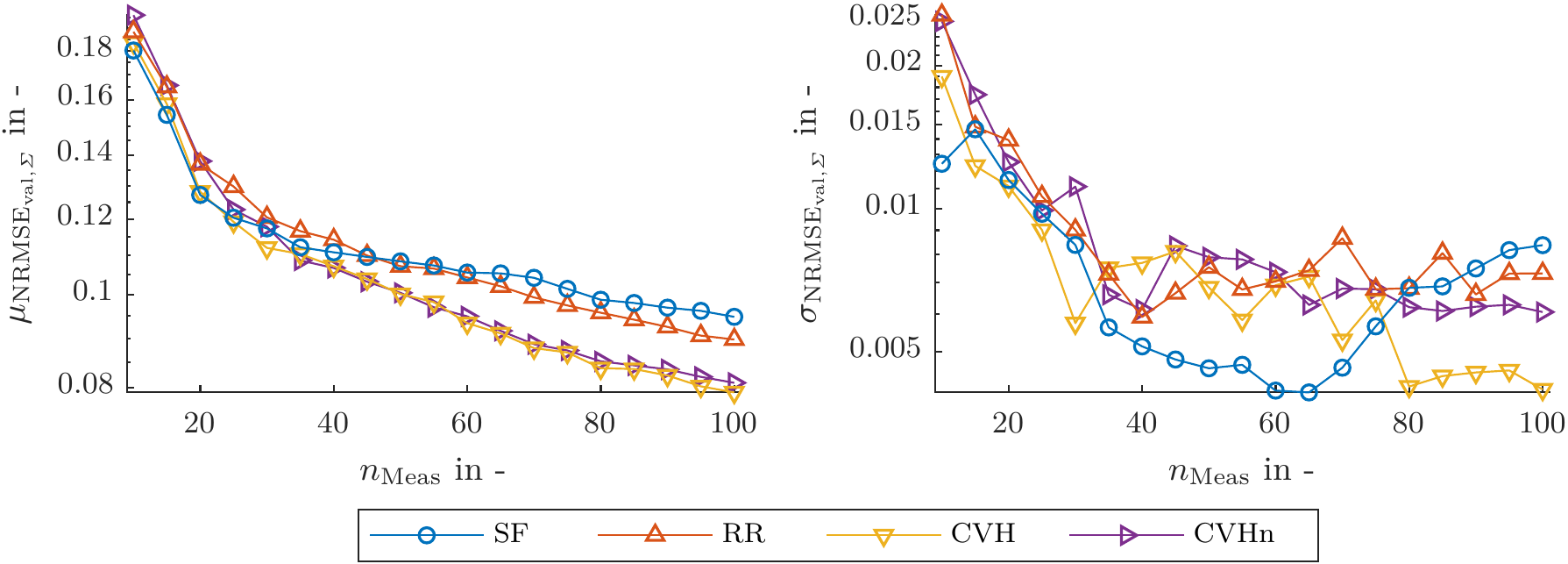}
	\caption{Mean $\mu$ and standard deviation $\sigma$ of the $\NRMSEF_{\text{val,}\varSigma}$ of setup~2 over the number of measurements \nMeas{}.}
	\label{fig:differentComplexity}	
\end{figure*}

Figure~\ref{fig:differentComplexity} shows the performances of those strategies on setup~2.
The benefits of considering \CVten{} for determining the leading output is clear at this point.
CVHn performs similarly well as CVH; both reduce the number of points by 35\% and 40\% compared to SF and by 30\% each compared to RR.
This is accomplished with a lower standard deviation over the runs.
CVH has a slightly lower mean value $\mu_{\NRMSEF_{\text{val,}\varSigma}}$ than CVHn over all runs.
However, this difference is not statistically significant.
This shows that the newly presented strategy CVHn performs as well as the already shown CVH in a setup were the latter showed to be very effective in the earlier publication, see \cite{prochaska_active_2020}.

\begin{figure*}
	\centering
	\includegraphics{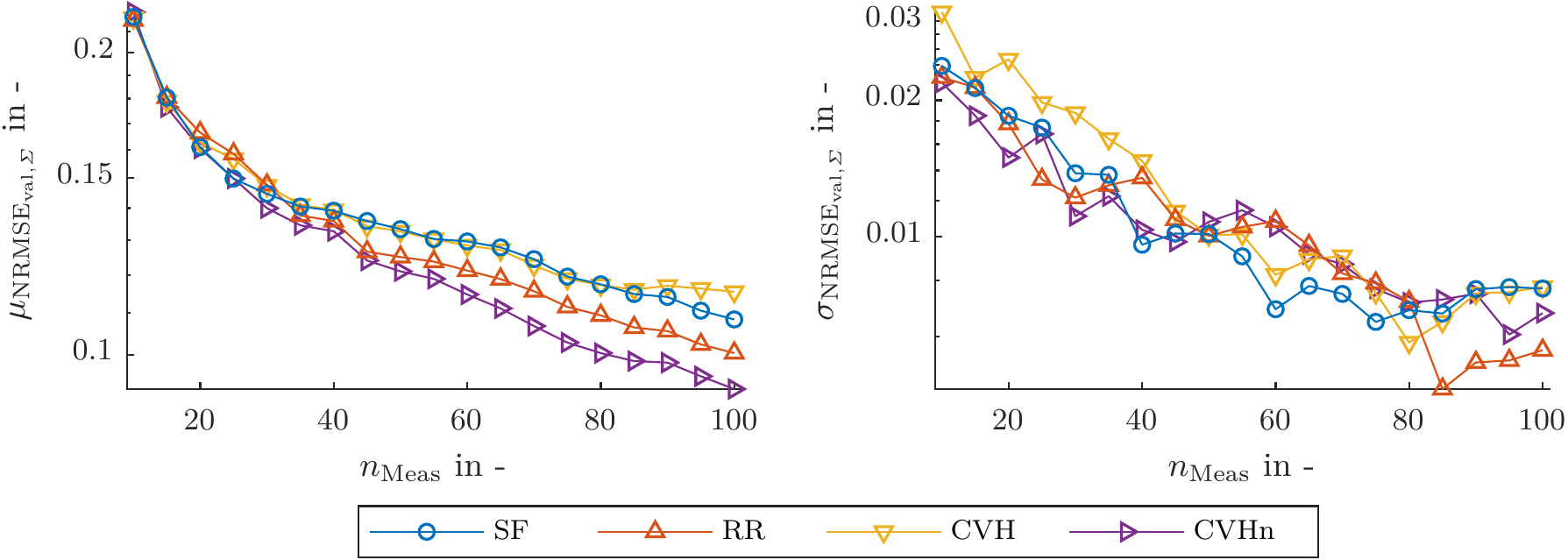}
	\caption{Mean $\mu$ and standard deviation $\sigma$ of the $\NRMSEF_{\text{val,}\varSigma}$ of setup~3 over the number of measurements \nMeas{}.}
	\label{fig:differentNoise}	
\end{figure*}

In the same paper, setup~3 showed the limits of CVH. 
The strategy often puts the noisy output in the lead because of its high \CVten{}.
Figure~\ref{fig:differentNoise} shows the evolution that takes place from CVH to CVHn.
While CVH is not even able to outperform the passive SF in the end, the robust heuristic RR reaches the end value of SF after 85\% of the measurements. 
The consideration of the models' estimated noise $\stdN{}$for normalization of \CVten{} reduces the drawbacks of CVH.
After $\nMeasF\geq60$, CVHn already outperforms CVH.
Even in this setup, CVHn reaches the end value of SF and RR after 70\% and 80\% of the measurements.

The results convincingly show the advantages of the new CVHn strategy.
Furthermore, the limitations of CVH in the presence of noise compared to CVHn are very apparent.

\section{Conclusion}
\label{sec:conclusion}
This paper introduces an improvement for the CVH active output selection strategy, which was presented by \cite{prochaska_active_2020}.
Active output selection strategies identify multiple process outputs with the same input dimensions.
Depending on the strategy, different outputs are selected as leading and therefore actively querying measurements.
CVH uses the 10-fold cross-validation error to determine the leading model. This strategy showed drawbacks in certain scenarios.
The new strategy CVHn normalizes the 10-fold cross-validation error to increase the informative content and takes into account a noise estimate of each process output.

The paper analyzes CVHn and compares it to CVH as well as two other baseline strategies.
Three toy examples are used to test said strategies in a reproducible but noisy environment.
Each of those toy examples addresses different difficulties in an active output selection task.
The results show that active learning strategies outperform the passive space-filling baseline strategy in most cases.
In one of the presented examples, CVHn performs similar to the existing CVH, but there is no statistical significance.
In any other example, CVHn outperforms all other strategies.
In a best case scenario, the number of points is reduced by 10\% at least and 60\% at best.
Specific weaknesses of CVH concerning the robustness to noise could be reduced with almost no compromise.

These results confirm the effectiveness of CVHn.
However, this paper is lacking results of a benchmark data set or real-world applications examples.
Therefore, further research must focus on testing the strategy in a real-world environment and creating a benchmark data set fitting to the specific application.
Moreover, this paper focuses on a very simple active learning strategy for each model. 
Implementation of more advanced active learning methods could further decrease the needed number of points. 
The AOS strategy, which was presented here, is very likely to benefit from such an implementation.

\bibliography{VO4-normalized_learner}             

\end{document}